\documentclass[11pt,a4paper]{article}
\usepackage[hyperref]{naaclhlt2018}
\usepackage{times}
\usepackage{latexsym}
\usepackage{macros}
\usepackage[hyphenbreaks]{breakurl}

\aclfinalcopy 

\newcommand\wtq{{\textsc{WikiTableQuestions}}}

\title{It was the training data pruning too!}

\author{
  $\mbox{Pramod Kaushik Mudrakarta}^1$,
  $\mbox{Ankur Taly}^2$,
  $\mbox{Mukund Sundararajan}^2$, and
  $\mbox{Kedar Dhamdhere}^2$\\
    $^1$The University of Chicago, ~~$^2$Google Research  \\
    {\tt pramodkm@uchicago.edu}, ~~{\tt \{ataly, mukunds, kedar\}@google.com}\\}

\date{}

\begin{document}
\maketitle
\begin{abstract}
We study the current best model~\cite{KDG17} (KDG) for question answering on
tabular data evaluated over the \wtq\ dataset.
Previous ablation studies
performed against this model attributed the model's performance to certain aspects
of its architecture. In this paper, we find that the model's performance also
crucially depends on a certain pruning of the data used to train the model.
Disabling the
pruning step drops the accuracy of the model from 43.3\% to
36.3\%.
The large impact on the performance of the KDG model
suggests that the pruning may be a useful pre-processing step in
training other semantic parsers as well.
\end{abstract}

\section{Introduction}
Question answering on tabular data is an important problem in natural
language processing. Recently, a number of systems have been proposed
for solving the problem using the \wtq\ dataset~\cite{PL15} (henceforth
called WTQ). This
dataset consists of triples of the form $\la$question, table,
answer$\ra$ where the tables are scraped from Wikipedia and
questions and answers are gathered via crowdsourcing. The dataset is
quite challenging, with the current best model~\cite{KDG17} (henceforth
called KDG) achieving a single model\footnote{A 5-model ensemble
  achieves a higher accuracy of 45.9\%.  We choose to work with a
  single model as it makes it easier for us to perform our analysis.}
accuracy of only 43.3\%\footnote{All quoted accuracy numbers are based
  on our training of the KDG model using the configuration and
  instructions provided at:
  \burl{https://github.com/allenai/pnp/tree/wikitables2/experiments/wikitables}}
. This is nonetheless a significant improvement compared to the 34.8\% accuracy
achieved by the previous best single model~\cite{HOP17}.

We sought to analyze the source of the improvement achieved by the KDG model.
The KDG paper claims that the improvement stems from certain aspects
of the model architecture.

In this paper, we find that a large part of the improvement also stems from
a certain pruning of the data used to train the model. 
The KDG system generates its training data using an algorithm proposed
by~\citet{PL16}. This algorithm applies a pruning step
(discussed in Section~\ref{subsec:pruning}) to eliminate spurious training
data entries.
We find that without this pruning of the training data, accuracy of the KDG
model drops to 36.3\%.
We consider this an important finding as the pruning step
not only accounts for a large fraction of the improvement in the
state-of-the-art KDG model but may also be relevant to training
other models.
In what follows, we briefly discuss the pruning algorithm,
how we identified its importance for the KDG model, and its
relevance to further work.

\section{KDG Training Data}
The KDG system operates by translating a natural language question and a table
to a \emph{logical form} in Lambda-DCS~\cite{L13}. A logical form is an executable
formal expression capturing the question's meaning. It is executed on the table
to obtain the final answer. 

The translation to logical forms is carried out by a deep neural network, also called
a \emph{neural semantic parser}.
Training the network requires a dataset
of questions mapped to one or more logical forms. The WTQ dataset only contains
the correct answer label for question-table instances. To obtain the desired
training data, the KDG system enumerates consistent logical form candidates 
for each $\la q, t, a\ra$ triple in the WTQ dataset, i.e., it enumerates
all logical forms
that lead to the correct answer $a$ on the given table $t$. For this, it relies
on the dynamic programming algorithm of~\citet{PL16}. This algorithm is called
\emph{dynamic programming on denotations} (DPD).

\subsection{Pruning algorithm}\label{subsec:pruning}
A key challenge in generating consistent logical forms
is that many of them are \emph{spurious}, i.e., they do not
represent the question's meaning. For instance, a spurious logical
form for the question ``which country won the highest number of gold
medals'' would be one which simply selects the country in the first
row of the table. This logical form leads to the correct answer only
because countries in the table happen to be sorted in descending order.

\citet{PL16} propose a separate algorithm for pruning out spurious logical
forms using fictitious tables. Specifically, for each question-table instance
in the dataset, fictitious
tables are generated\footnote{Fictitious tables are generated by applying perturbations,
  such as shuffling columns, to the original table.
We refer to~\cite{PL16} for more details.}, and answers are
crowdsourced on them. A logical form that fails to obtain the correct answer
on any fictitious table is filtered out. The paper presents an
analysis over 300 questions revealing that the algorithm eliminated
92.1\% of the spurious logical forms.

\subsection{Importance of pruning for KDG}
The KDG system relies on the DPD algorithm for generating consistent
logical form candidates for training. It does not explicitly prescribe pruning out
spurious logical forms candidates before training.
Since this training set contains spurious logical forms, we expected the model
to also sometimes predict spurious logical forms. 

However, we were somewhat surprised to find that the logical forms
predicted by the KDG model were largely non-spurious. 
We then examined the logical form candidates\footnote{
Available at:
\burl{https://cs.stanford.edu/~ppasupat/research/h-strict-all-matching-lfs.tar.gz}}
that the KDG model
was trained on. 
Through personal communication with Panupong Pasupat, we
learned that all of these candidates had been pruned using the
algorithm mentioned in Section~\ref{subsec:pruning}.

We trained the KDG model on unpruned logical form 
candidates\footnote{Available at:  \burl{https://nlp.stanford.edu/software/sempre/wikitable/dpd/h-strict-dump-combined.tar.gz}}
generated using the DPD algorithm,
and found its accuracy to drop to 36.3\% (from 43.3\%); all
configuring parameters were left unchanged\footnote{It is possible
  that the model accuracy may slightly increase with some
  hyper-parameter tuning but we do not expect it to wipe out the large
  drop resulting from disabling the pruning.}.
This implies that pruning out spurious logical forms before
training is necessary for the performance improvement
achieved by the KDG model.

\subsection{Directions for further work}
\citet{PL16} claimed ``the pruned set of logical
forms would provide a stronger supervision signal for training a semantic parser''.
This paper provides empirical evidence in support of this claim.
We further believe that the pruning algorithm may also be valuable to models
that score logical forms. Such scoring models are typically used
by grammar-based semantic parsers such as the one in~\cite{PL15}.
Using the pruning algorithm, the scoring model can be trained
to down-score spurious logical forms.
Similarly, neural semantic parsers trained using reinforcement learning
may use the pruning algorithm to only assign rewards to non-spurious logical
forms.

The original WTQ dataset may also be extended with the fictitious
tables used by the pruning algorithm.
This means that for each $\la q, t, a\ra$ triple in the original dataset,
we would add additional triples $\la q, t_1, a_1\ra,
\la q, t_2, a_2\ra,\ldots$ where $t_1, t_2,\ldots$ are the fictitious tables
and $a_1, a_2\ldots$ are the corresponding
answers to the question $q$ on those tables. Such training data augmentation
may improve the performance of neural networks that are directly trained
over the WTQ dataset, such as ~\cite{NLAMA17}.
The presence of fictitious tables in the training
set may help these networks to generalize better, especially on tables that are outside
the original WTQ training set.


\section{Discussion}
\citet{KDG17} present several ablation studies to identify the
sources of the performance improvement achieved by the KDG model.
These studies comprehensively cover novel aspects of the model
architecture. On the training side, the studies only vary the number
of logical forms per question in the training dataset.
Pruning of the logical forms was not considered.
This may have happened inadvertently as the KDG system
may have downloaded the logical forms dataset made available by Pasupat et al.
without noticing that it had been pruned out\footnote{This was initially the
  case for us until we learned from Panupong Pasupat that
  the dataset had been pruned.}.

We note that our finding implies that
pruning out spurious logical forms before training is an important
factor in the performance improvement achieved by the KDG model.
It does \emph{not} imply that pruning is the only important
factor. The architectural innovations are essential for the performance
improvement too.

In light of our finding, we would like to emphasize that the
performance of a machine learning system depends on several factors
such as the model architecture, training algorithm, input
pre-processing, hyper-parameter settings, etc.
As~\citet{KL88}  point out, attributing
improvements in performance to the individual factors is a valuable
exercise in understanding the system, and generating ideas for
improving it and other systems.  
In performing these
attributions, it is important to consider all factors that may be
relevant to the system's performance.

\section*{Acknowledgments}
We would like to thank Panupong Pasupat for helpful discussions on the pruning
algorithm, and for providing us with the unpruned logical form candidates. We would like to thank Pradeep Dasigi for helping us train the KDG model.
\bibliography{paper}
\bibliographystyle{acl_natbib}

\appendix


\end{document}